%% file: main.tex
\documentclass[letterpaper, 10 pt, conference]{ieeeconf}
\IEEEoverridecommandlockouts
\usepackage{amsmath,amssymb}
\usepackage{newtxtext,newtxmath}
\usepackage{booktabs}
\usepackage{graphicx}
\usepackage{algorithm}
\usepackage{algpseudocode}
\usepackage{xcolor}
\usepackage[colorlinks=true,linkcolor=blue!55!black,citecolor=blue!55!black,urlcolor=blue!55!black]{hyperref}

\providecommand{\citep}{\cite}
\providecommand{\citet}{\cite}
\newcommand{\dmax}{\Delta_{\max}}

\title{\LARGE \bf Faster Imitation by Re-timing Demonstrations Within the Controller Limit}
\author{Zongtan Li$^{1}$, Tianming Liu$^{1}$, Ye Zhao$^{2}$\\
$^{1}$University of Georgia, Athens, GA, USA\\
$^{2}$Georgia Institute of Technology, Atlanta, GA, USA}

\begin{document}
\maketitle
\thispagestyle{empty}
\pagestyle{empty}

\begin{abstract}
Imitation policies inherit the operator's pace, and re-timing the demonstrations to run faster works only until success collapses at two to three times demonstration speed. Recent methods attribute that collapse to distribution shift and answer it with finer rules for which segments may be sped up; we show it has a mechanical cause that segment selection cannot address. A low-level controller realises only a bounded displacement per control period, so merging consecutive commands produces sums outside that bound: training pairs that promise a transition the plant cannot deliver. The share of such pairs in the re-timed data, which we call the \emph{exceed share}, tracks where the retrained policy collapses. The rule that follows has a single parameter: merge greedily while the sum stays realisable. Every command is then realisable by construction, the demonstrated geometry is preserved, and the speed-up lands only where the demonstrator left slack. On four simulated \textsc{robomimic} tasks, 200 held-out seeds per cell, a merge cap of two is at or above the frozen policy's success on all four, by 14 points on \textsc{Square}, and shortens the successful episodes. At matched tempo, uniform re-timing loses 9 and 16 points on \textsc{Can} and \textsc{ToolHang}, is within noise on the other two, and reaches zero on the hardest at $2\times$. Placement also matters: no shuffled, entropy-guided or speed-conditioned alternative beats the rule on three of the four tasks, by margins the exceed share alone does not explain; on the fourth, at a tempo the task absorbs, every placement lands in the same band. The same principle carries to a simulated humanoid whole-body tracker: re-timing 429 reference motions within its velocity envelope succeeds on 71\% of clips at $2.3\times$, against 51\% for uniform time scaling at $2\times$.
\end{abstract}

\section{Introduction}
A robot that learns from human teleoperation learns the operator's pace, and that pace is slow. On the \textsc{robomimic} datasets human operators use a third to two thirds of the displacement the controller allows, while a machine-generated policy on the same tasks uses nearly all of it. The slowness is uniform rather than concentrated in the hard parts: the operators' speed far from the object is within $0.88$ to $1.33$ of their speed near it. For a deployed robot the difference is throughput left on the table.

The obvious remedy is to re-time the demonstrations and retrain, and it works up to a point. Recent methods reach two to three times demonstration speed before success collapses~\citep{demospeedup2025,espada2025,volt2026,tempovla2026}. Most answer it with a finer rule about which segments may be sped up: the high-entropy ones under a generative proxy policy, the ones a vision-language model judges non-critical, or the ones a speed-conditioned policy is asked to run fast. TempoVLA is more specific and names the low-level controller's tracking bandwidth~\citep{tempovla2026}. What none of them measures is the quantity the shift is a shift in: a re-timed demonstration is a set of promises about transitions, and whether the plant keeps them is checkable on the data before any rollout.

We measure it, and the answer removes the premise those methods are built on. The controller realises a bounded displacement per control period. Merging $f$ consecutive commands produces sums that exceed the bound, and each such training pair tells the policy that a displacement will be realised which the plant will truncate. The successor state the policy learns to expect is not the one it reaches. This is a covariate shift built into the dataset before any rollout, and the share of such pairs, which we call the \emph{exceed share}, orders the collapse: on \textsc{Can} the uniform factors $1.5$, $2$ and $3\times$ give exceed shares of $0.30$, $0.60$ and $0.72$ and retrained success of $87\%$, $60\%$ and $1\%$.

The diagnosis leaves no freedom in the method. If the cliff is produced by training pairs the plant cannot honour, an accelerator must produce none, and the only remaining question is how far the data can be compressed subject to that. We merge consecutive commands greedily while their sum stays inside the realisable set, up to a cap $f_{\max}$. No merged command is clipped, every group boundary is a demonstrated observation, and where the speed-up lands is not ours to choose: it is decided by how much slack the demonstrator left at each step. The tempo this reaches, $1.25$ to $1.89\times$ on human demonstrations, is lower than the numbers the literature reports before its collapse; that is the bound rather than a tuning failure, and on machine-generated demonstrations that already use $95\%$ of the limit the rule merges almost nothing and returns the nominal policy.

Our contributions are:
\begin{itemize}
\item A mechanical account of the cliff in faster-than-demonstration imitation: the bounded displacement a controller realises per period, whose violation share in the re-timed training data tracks where retraining collapses, on a manipulator and on a humanoid tracker (\S\ref{sec:mech}).
\item Within-limit re-timing, the one-parameter rule that follows. At a merge cap of two it matches or exceeds the frozen policy's success while shortening episodes on all four \textsc{robomimic} tasks, including the two hardest, against a uniform baseline that loses up to 16 points at matched tempo and reaches zero on \textsc{ToolHang} (\S\ref{sec:exp}).
\item The consequence that choosing which segments to speed up is a question about the transition each merge produces, not about which segments are semantically critical. We test this against a matched-tempo shuffled control, an entropy-guided baseline, a speed-conditioned baseline, and a task where the limit does not bind (\S\ref{sec:controls}).
\end{itemize}
All experiments are in simulation and primarily use state observations.

\section{Related Work}
\paragraph{Faster-than-demonstration execution} SAIL~\citep{sail2025} modulates speed at execution with a high-gain tracker and notes that faster tracking pushes the policy into unfamiliar states. DemoSpeedup, ESPADA and VOLT segment demonstrations by action entropy, by vision-language semantics and by video labels, downsample the segments judged safe, and retrain~\citep{demospeedup2025,espada2025,volt2026}. TempoVLA conditions the policy on a speed scalar~\citep{tempovla2026} and AutoSpeed learns stage-adaptive speed~\citep{autospeed2026}, while Speedup Patch and PACS act on a frozen policy's actions at execution~\citep{sup2026,pacs2026}. Several report the practical ceiling as the controller's tracking bandwidth~\citep{tempovla2026,bspline2026}. We locate the ceiling one level deeper, in the displacement realised per command, and respect it in the data rather than discovering it at execution.

\paragraph{Time-optimal re-timing under actuator limits} Time-optimal path parameterisation re-times a fixed geometric path so that velocity, acceleration or torque bounds are met~\citep{bobrow1985,kunz2012,toppra2018}, and RACE applies it to each predicted chunk at execution~\citep{race2026}. Our rule is the same operation applied to a demonstration before training: the path is the demonstrated one, the limit is what the controller realises in a period, and the object kept feasible is the training pair rather than the executed trajectory.

\paragraph{Covariate shift in imitation} Compounding error is classical~\citep{ross2010,ross2011dagger} and its remedies act at collection or through interaction~\citep{laskey2017dart}. Spencer et al.~\citep{spencer2021} separate a bounded-density-ratio regime from an unbounded one and Block et al.~\citep{block2023} tie imitation guarantees to the controller's stability constants. Our mechanism names the part of the pipeline that produces the perturbation and removes it from the data.

\section{The Controller Limit as the Cause}
\label{sec:mech}
\subsection{Commands, realisable set, exceed share}
A policy maps an observation $o_t$ to a chunk of actions and a low-level controller executes the first $k$ before replanning. A demonstration is a sequence $(o_t,a_t)_{t=1}^{n}$ recorded at the control period $\tau$. A command has \emph{additive} channels $\delta_t\in\mathbb{R}^d$ that the controller integrates and \emph{setpoint} channels $\sigma_t$ that an actuator moves toward on its own time scale, such as a gripper state. Over one period the plant realises additive commands only inside a \emph{realisable set} $\mathcal{R}$:
\begin{equation}
\begin{split}
&\mathcal{R}=\{\delta:\|\delta\|_\infty\le\dmax\},\quad
\mathcal{R}=\{\delta:|\delta_j|\le\bar v_j\tau\},\\
&\mathcal{R}=\lambda\,\mathcal{R}_{\mathrm{doc}}
\end{split}
\label{eq:rsets}
\end{equation}
for, respectively, the per-axis clamp of the operational-space controller used here, the per-joint velocity envelope of a whole-body tracker, and a plant whose effective limit is a fraction $\lambda$ of its documented one. A command outside $\mathcal{R}$ is truncated by the controller. This form is first-order: it constrains one command in isolation. On a plant whose binding budget is acceleration rather than displacement, feasibility couples consecutive commands and the test must also bound their difference; \S\ref{sec:limits} returns to this.

Re-timing partitions each demonstration into groups $G_1,\dots,G_K$ of consecutive steps; group $G$ becomes one training pair with the observation at its start, the summed increment $\sum_{t\in G}\delta_t$, and the setpoint at its end. The re-timed dataset has
\begin{equation}
\rho=\frac{1}{K}\sum_{k}\mathbf{1}\Big\{\sum_{t\in G_k}\delta_t\notin\mathcal{R}\Big\},\qquad \mathrm{tempo}=\frac{n}{K}.
\label{eq:exceed}
\end{equation}
When $\rho=0$, executing a merged command from its group's start moves the plant by the displacement the demonstrator realised over the group, and the retrained policy's expected and realised successors agree up to the tracking error the demonstration already had. When $\rho>0$, each such pair falls short by the part of its sum outside $\mathcal{R}$, and under a per-axis clamp its direction changes as well.

\subsection{The shortfall as a threshold effect}
The consequence of a shortfall is not proportional to its size. We ran episodes at nominal speed and re-ran them from the same seed with a random multiplier over one random 16-step window, recording the window's own path deviation with the episode outcome, about 1000 labels per task and policy. A two-level step in $\log d_w$ fits better than a logistic slope on every task and both policy classes, by more than its one extra parameter accounts for on five of the six fits: success stays near nominal up to a threshold of $40$ to $90$\,mm, then falls to a flat floor of $15$ to $40\%$ however far the window strays. Compounding error predicts continued decay; leaving the training support with a fixed chance of recovery predicts a floor, and the data show a floor.

An accelerator should therefore drive $\rho$ to zero rather than make the average shortfall small. Uniform re-timing by $f$ fixes $|G|=f$ and accepts whatever $\rho$ results. Within-limit re-timing fixes $\rho=0$ and accepts whatever tempo results.

\subsection{Absorption of an execution-time multiplier}
The same limit explains why running a frozen policy faster buys so little. Resolved by phase, a $1.25\times$ multiplier on the \textsc{Can} policy is clipped on three quarters of the transport steps and four fifths of the carry steps, against $3\%$ of the approach steps: the policy already commands near-maximal displacements where it moves fast. The multiplier is absorbed there and acts in full only in the slow approach and grasp, which is exactly where a faster pace puts the policy in states it never visited. Under a $2\times$ multiplier on \textsc{Can} the successful episodes shorten by $13\%$ while success drops 26 points.

\section{Method}
\label{sec:method}
Algorithm~\ref{alg:retime} builds the groups greedily and left-maximally, extending a group while
\begin{equation}
\begin{split}
\mathrm{Feasible}(G)\iff\ &\textstyle\sum_{t\in G}\delta_t\in\mathcal{R}\ \wedge\ |G|\le f_{\max}\\
&\wedge\ \sigma\ \text{settled on}\ G\ \text{(optional)} .
\end{split}
\label{eq:feasible}
\end{equation}
A fresh policy is trained on the re-timed pairs with the same recipe as the frozen one and executed at native speed. The algorithm is linear in $n$, needs no model of the policy, and has one parameter.

\begin{algorithm}[t]
\caption{Within-limit re-timing of one demonstration}
\label{alg:retime}
\begin{algorithmic}[1]
\Require commands $(\delta_t,\sigma_t)_{t=1}^n$, observations $(o_t)_{t=1}^n$, set $\mathcal{R}$, cap $f_{\max}$
\State $t\gets1$, $\mathcal{D}\gets\emptyset$
\While{$t\le n$}
  \State $e\gets t$
  \While{$e<n$ and $\mathrm{Feasible}(\{t,\dots,e+1\})$}
    \State $e\gets e+1$
  \EndWhile
  \State $\mathcal{D}\gets\mathcal{D}\cup\{(o_t,\ (\textstyle\sum_{i=t}^{e}\delta_i,\ \sigma_e))\}$, $\ t\gets e+1$
\EndWhile
\State \Return $\mathcal{D}$ \Comment{$\rho(\mathcal{D})=0$ by construction}
\end{algorithmic}
\end{algorithm}

Provided every demonstrated command is itself realisable, $\rho=0$ by construction: a group of one step is accepted untested, and nothing shorter exists to repair it. Group boundaries are demonstrated observations, though the motion between two of them is the chord and not the demonstrated polyline. The tempo is at most $f_{\max}$ and, absent sign cancellation, at most $1/\max_j\bar u_j$, the inverse of the largest per-axis mean utilisation: $1.77\times$ on \textsc{Can} against a measured ceiling of $1.42\times$ as $f_{\max}$ grows, $1.25\times$ at the cap of two used in the main result, and $1.14\times$ on machine-generated \textsc{Can}. Fewest groups subject to a per-group constraint is a shortest path over step boundaries, solvable exactly in $O(nf_{\max})$; greedy attains it where feasibility is hereditary, as for the arc form, and is within $0.1\%$ of it under the chord form used here, which is not. That audit is of the command-constrained problem, not of a transition-constrained one. Write the \emph{replay gap} of a group for the distance between the state the plant reaches under the group's original command sequence and the state it reaches under the single merged command, both started from the same state. Measuring it on every candidate group on \textsc{Can} settles the second audit as well: the clamp already excludes almost every expensive merge, the median gap being $21$\,mm outside it against $0.0$\,mm inside, so the admissible set is nearly flat in the quantity the command test stands for. Solving the fixed-tempo problem exactly on the measured gaps lowers their sum by under $5\%$, changes $6\%$ of the group boundaries, and leaves the retrained success unchanged, $96.2\%$ against $95.7\%$ over three seeds and $600$ paired episodes. The rule is therefore close to optimal within its objective class, and what headroom remains lies in trajectory-level objectives rather than in better per-group grouping.

\paragraph{The gripper is a second actuator} A setpoint channel is realisable over a group when it does not change faster than its actuator can follow, so the general rule holds it fixed within a group. On \textsc{Lift} and \textsc{Can} this changes nothing measurable, because the simulated gripper closes within a few periods. At the larger cap on the precision tasks it is not optional: merging through the closing motion costs 26 points for a flow-matching policy on \textsc{Square} and ten points on \textsc{ToolHang}. At $f_{\max}=2$ on \textsc{Square} the hold is not needed and costs tempo, so we report that task without it and \textsc{ToolHang} with it. The hold binds on any plant whose gripper takes more than one control period to move, which is why the general form rather than the position clamp is what goes to hardware; whether a given gripper does is an empirical question we do not settle here.

\paragraph{Identifying the limit} The limit need not equal the command clamp: a stiffer controller overshoots on large steps and stops realising a command well below it. Two instruments find it without touching the policy: an open-loop replay of the re-timed demonstrations, which fails on the data of any merge that exceeds it, and a calibration that picks the largest merge limit whose retrained success stays nominal. On a Panda with twice the stock gain, merging under the clamp loses 24 points while merging under half of it returns nominal success. A limit assumed too small costs tempo; one assumed too large costs success.

\paragraph{Continuous form} When the demonstration is a reference a controller tracks, merging is the same as playing time faster and the rule becomes a local time scaling $s_t=\min(k_{\max},\min_j \bar v_j/|v_j(t)|)$, the form used on the humanoid.

\section{Experiments}
\label{sec:exp}
\subsection{Protocol}
The manipulation experiments use the \textsc{robomimic}~\citep{robomimic2021} \textsc{Lift}, \textsc{PickPlaceCan}, \textsc{NutAssemblySquare} and \textsc{ToolHang} tasks with proficient-human (PH), multi-human (MH) and machine-generated (MG) demonstrations, in \textsc{robosuite} on a Panda arm with the stock operational-space controller at 20\,Hz and $\dmax=0.05$\,m per period. Observations are end-effector pose, gripper opening and object pose. Two policy classes are trained: an MLP that regresses an 8-step action chunk, and a conditional flow-matching policy~\citep{lipman2023flow} of the class used by $\pi_0$-style VLAs~\citep{pi0_2024}. All policies execute 4 steps per chunk and replan, and every retrained policy uses the same network, data and recipe as the frozen one. Arms are evaluated on seeds 1000--1199, disjoint from every seed used elsewhere. Cells give success @ time to success, the mean steps of the successful episodes. With 200 episodes the $95\%$ interval on success is about $\pm4$ points at $90\%$ and $\pm7$ at $60\%$. Every arm runs on the same seeds, so comparisons are paired and tested with an exact McNemar test. Where a cell was trained at three seeds, the table gives the mean and the seed range.

\begin{figure*}[t]
\centering\includegraphics[width=0.94\linewidth]{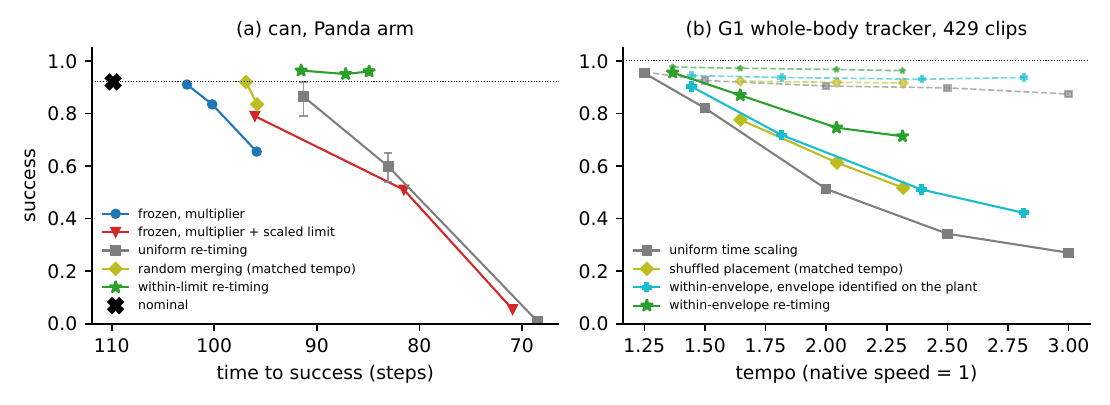}
\caption{(a) Success against time to success on \textsc{Can} for every route, 200 held-out seeds; up and to the right is better, and the cross is the frozen policy at nominal speed. Error bars give the range over three training seeds where those exist, for uniform $1.5\times$ and $2\times$ and for within-limit $f_{\max}=2$ and $4$; the remaining points are a single training seed. (b) Success against tempo for the G1 whole-body tracker on 429 reference clips; dashed lines are the share of clips without a fall.}
\label{fig:routes}
\end{figure*}

\begin{table*}[t]
\centering\footnotesize
\setlength{\tabcolsep}{4pt}\input{tables/main}
\caption{Every route on the four tasks, 200 held-out seeds per cell. Cells give success @ time to success, with the training-seed range as a subscript and the tempo of the re-timed data in brackets. Uniform $3\times$ reaches $1\%$ on \textsc{Can} and $0\%$ on \textsc{ToolHang} and is omitted.}
\label{tab:main}
\end{table*}

\begin{table*}[t]
\centering\footnotesize
\setlength{\tabcolsep}{4pt}\input{tables/controls}
\caption{Placement at comparable tempo. The shuffled row produces the same number of training pairs from the same demonstrations as the within-limit row and differs only in which steps are merged; the entropy-guided and speed-conditioned rows run at the tempo shown in brackets. The entropy row merges the low-scoring half, inverting the direction DemoSpeedup specifies; the speed-conditioned row is that policy's best cell. The within-limit row is the same replicate set as Table~\ref{tab:main}; the control arms are a single training seed.}
\label{tab:controls}
\end{table*}

\begin{table*}[t]
\centering\footnotesize
\setlength{\tabcolsep}{4pt}\input{tables/phase}
\caption{Placement at matched tempo \emph{and} matched exceed share on \textsc{Can}, over three training seeds. All four arms carry the same tempo and the same per-group exceed share by construction, and differ only in which phase the surviving exceedances are left in. The replay gap $\hat\eta$ is the endpoint error over the four-group execution window, measured on the partitions alone and involving no policy.}
\label{tab:phase}
\end{table*}

\subsection{Main result}
\begin{figure}[t]
\centering\includegraphics[width=0.98\linewidth]{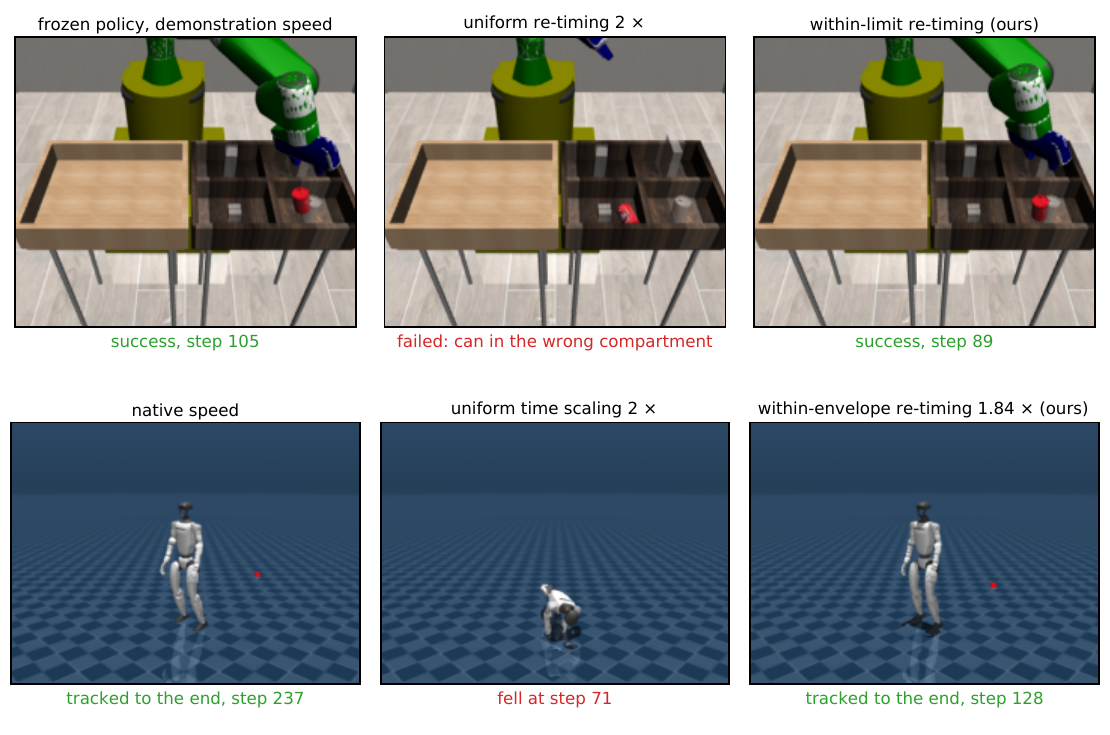}
\caption{Final frames of one episode per arm. Top, \textsc{Can}: uniform re-timing at $2\times$ puts the can in the wrong compartment, while the frozen policy and within-limit re-timing both succeed, the latter in fewer steps. Bottom, G1 on a jog with a 180-degree turn: the tracker falls under uniform time scaling at $2\times$ and follows the within-envelope reference at $1.84\times$ to the end.}
\label{fig:keyframes}
\end{figure}

Table~\ref{tab:main} and Fig.~\ref{fig:routes} give every route, and Fig.~\ref{fig:keyframes} shows one episode each. Within-limit re-timing at $f_{\max}=2$ is at or above the frozen policy on all four tasks while shortening the successful episodes, and uniform re-timing at a comparable tempo is below it on three, well below on two. \textsc{Can} is the clearest case, where the rule holds $96\%$ against $60\%$ for uniform $2\times$ at a similar realised completion time, on every one of the nine seed pairings (exact McNemar, $p<10^{-15}$). \textsc{ToolHang} is the hardest, and the task where the gap is widest: the rule wins against uniform on every one of the nine seed pairings ($p\le0.04$) while more than halving the length of a successful episode, and uniform $2\times$ and $3\times$ solve at most one episode in 200 there.

The direction of the clipped command is not what matters: scaling a merged sum back into the limit so its direction is preserved is worse still, $48\%$ on \textsc{Can} at $2\times$, because it realises even less of the promised displacement. \textsc{Lift} has no precision phase and its transport sits far below the limit, so nothing collapses there.

\subsection{Placement at matched tempo}
\label{sec:controls}
Table~\ref{tab:controls} compares placements at comparable tempo. On \textsc{Lift}, \textsc{Can} and \textsc{ToolHang} no alternative placement beats the within-limit rule, outright on \textsc{Can} and \textsc{ToolHang} and on success alone on \textsc{Lift}, where the shuffled arm ties at $99\%$ but takes longer, and each leaves unrealisable commands behind: the shuffled control clips $13$ to $21\%$ of the data, and entropy-guided merging of the low-scoring half $4$ to $20\%$. Merging the high-scoring half, the direction DemoSpeedup specifies, is worse than either, because high entropy marks the fast transport where the demonstrator has no slack. That arm is also the one which shows the exceed share to be insufficient on its own: it matches uniform $1.5\times$ on both the frequency and the severity of its violations to within $3\%$, runs at a lower tempo, and still loses 34 points. Speed conditioning stays below the within-limit rule at every requested speed on \textsc{Lift}, \textsc{Can} and \textsc{ToolHang}, consistent with its pooling the unrealisable pairs of every speed rather than separating them. Table~\ref{tab:phase} sharpens that insufficiency. Four re-timings of \textsc{Can} are matched by construction on tempo and on per-group exceed share and differ only in which phase the surviving exceedances fall in. They separate by up to nine points, mainly between the placement that spares the transport phase and the other three, and their replay gaps order the same way. Equal tempo and equal exceed share therefore do not fix performance. \textsc{Square} is the exception to the ordering above and is treated next.

The complement is \textsc{Square}, where at a tempo the task absorbs, placement mostly stops mattering. Uniform, shuffled, low-half entropy-guided and within-limit merging all reach $70$ to $76\%$ near a tempo of $1.5\times$ against the frozen policy's $62\%$, falling together to $59$ to $66\%$ at $2\times$; the high-half entropy arm is the exception at $53\%$. That budget is global in tempo, wherever the speed is placed. The limit still binds on $11$ to $18\%$ of the merged commands across the coinciding arms, so it is not the absence of violations that makes them agree. At the larger tempo the rule reaches on this task, the low-half entropy and speed-conditioned arms score higher at their own lower tempos, so the ordering of the other three tasks is not universal.

\subsection{Generality}
\begin{table}[t]
\centering\scriptsize
\setlength{\tabcolsep}{2pt}\input{tables/general}
\caption{Scope of the result. Cells give success @ time to success on 200 held-out seeds.}
\label{tab:general}
\end{table}

\begin{table}[t]
\centering\footnotesize
\setlength{\tabcolsep}{3pt}\input{tables/g1}
\caption{Time scaling of reference motions for the G1 humanoid under the released SONIC tracker, 429 clips. Success is no fall and a tracking error within $1.25\times$ the native-speed value.}
\label{tab:g1}
\end{table}

The mechanism belongs to the controller and the data, so the result should survive a change of policy, demonstrator and plant, and vanish where nothing exceeds the limit (Table~\ref{tab:general}). The flow-matching policy is more fragile under uniform re-timing than the MLP, $33\%$ at $2\times$ on \textsc{Can}, and recovers most of that loss, $90\%$ at a similar realised completion time. Mixed-skill demonstrations give a policy that stalls, and the speed-up helps it twice over: $55\%$ to $80\%$ on \textsc{Can} and $77\%$ to $96\%$ on \textsc{Lift}. Machine-generated demonstrations are the case the tempo bound predicts the rule cannot help: their commands use $95\%$ of the range, the rule merges almost nothing and returns the nominal policy, while uniform $2\times$ clips $91\%$ of its commands and falls from $66\%$ to $30\%$. The ordering survives a change in the number of demonstrations and in the chunk horizon, because the exceed share does: with 50 demonstrations, and with a chunk of 16 executed 8 at a time, the training exceed share of uniform $2\times$ stays near $0.60$ in every case.

\paragraph{A humanoid whole-body tracker} The same rule holds on a different morphology and a different kind of controller. The Unitree G1 with the released SONIC tracker~\citep{sonic2025} follows a reference motion with no explicit clamp. We take 13 clips released with the tracker and 515 retargeted from a motion-capture library, and keep the 429 the tracker follows at native speed without falling. The envelope $\bar v$ is the per-joint 99th percentile of reference velocities over those clips, and playing a clip $k$ times faster is the analogue of uniform re-timing. Table~\ref{tab:g1} shows the same picture as the arm: uniform time scaling falls from $95\%$ at $1.25\times$ to $51\%$ at $2\times$, while within-envelope re-timing holds $87\%$ at $1.65\times$ and $71\%$ at $2.32\times$, winning the per-clip paired comparison at every $k_{\max}$. Binned by exceed share, the uniform arms succeed on $97\%$ of clips whose exceed share is zero and $31\%$ above a share of $0.4$, and the share that fall rises from $2\%$ to $16\%$ along the same bins. A shuffled placement at matched tempo loses to the within-envelope one, so placement carries the gain here as on the arm. The tracker keeps a second budget of its own: its tracking error grows continuously with tempo, from $1.07\times$ native at $1.37\times$ to $1.21\times$ at $2.32\times$, which respecting the envelope does not remove.

\begin{table}[t]
\centering\footnotesize
\setlength{\tabcolsep}{3pt}\input{tables/vla}
\caption{A pretrained vision-language-action policy on \textsc{Can}. SmolVLA-base fine-tuned for 12k steps on two rendered $128\times128$ views and proprioception, with no object pose; three training seeds of 200 episodes each. Columns give tempo, training exceed share, per-seed success and the mean with time to success.}
\label{tab:vla}
\end{table}

\section{Limitations}
\label{sec:limits}
The realisable set is first-order. On a plant whose binding budget is joint acceleration, a dataset with displacement exceed share zero can still demand accelerations the plant will not deliver, and the natural repair is to bound the difference between consecutive merged commands as well as each command. The greedy rule cannot enforce such a bound: choosing only boundaries, it must accept untested the group of one that refusing an extension creates, and so violates its own bound the more often the tighter it is set, on $3.6\%$ of transitions at $0.58\,\dmax$ and $21\%$ at $0.25\,\dmax$, while the tempo falls from $1.31$ to $1.15$. That is a limitation of the rule and not of the constraint. Carrying the incoming group length in the state makes the coupled problem an exact shortest path in $O(nf_{\max}^2)$, and its solution improves on greedy in violations and tempo together on all three tasks; the violations it cannot remove track the demonstration's own consecutive-command variation, which groups of one reproduce. Neither partition changes task success on the stock or the stiff plant, so what the second-order case leaves open is which budget to write down rather than how to optimise one. A scalar bound in displacement space is also the wrong currency for a budget that lives in joint space, and the controller used here is damped enough that the proxy never binds.

All experiments here are in simulation with state observations and small policies. On a pretrained image-conditioned VLA (SmolVLA-base~\citep{smolvla2025}, three seeds) within-limit re-timing beats uniform $2\times$, $71\%$ against $49\%$, but both stay below the nominal policy's $87\%$ (Table~\ref{tab:vla}), so that experiment does not demonstrate a useful acceleration of the pretrained policy. The step thresholds are empirical and not derived from the policy or the controller. The method's gain is bounded by the slack of the demonstrations and by the limit it respects: $1.25$ to $1.89\times$ on human demonstrations, $1.10\times$ on machine-generated ones and $1.06\times$ on a stiff arm. Going beyond requires raising the limit at execution, which costs success on the contact tasks. On plants whose effective limit lies below the clamp the limit must be identified first. On the humanoid the retraining half of the pipeline is untested, since the tracker is fixed. The literature baselines are our reimplementations of the published operations rather than the original systems, which use vision-language segmentation we do not have.

\section{Conclusion}
In this paper, we proposed within-limit re-timing, a one-parameter rule that compresses a demonstration by merging consecutive commands only while their sum stays inside the displacement the controller realises in one period. The rule follows from a diagnosis: uniform re-timing writes commands the plant cannot execute into the training data, and the share of such commands tracks where retraining collapses. Across four manipulation tasks, two policy classes, three demonstration sources and a humanoid whole-body tracker, re-timing within the limit improves the trade-off between success and completion time over uniform re-timing at comparable speed, and the tempo it reaches is read from the slack the demonstrator left rather than chosen in advance. Matching the exceed share does not by itself equalise performance, so where compression is placed remains consequential beyond how much of it there is. Extending the constraint from a first-order displacement bound to controller state and contact, and validating the rule on physical hardware, are the natural next steps.

{\scriptsize
\bibliographystyle{IEEEtran}
\bibliography{refs}}

\end{document}

%% file: tables/main.tex
\begin{tabular}{lcccc}
\toprule
route & \textsc{Lift} & \textsc{Can} & \textsc{Square} & \textsc{ToolHang} \\
\midrule
frozen policy, nominal speed & 94\% @ 45 & 92\% @ 110 & 62\% @ 168 & 16\%$_{14\text{--}20}$ @ 536 \\
\midrule
uniform re-timing $1.5\times$ & 96\% @ 33 & 87\%$_{79\text{--}92}$ @ 91 & 72\% @ 131 & 8\%$_{6\text{--}10}$ @ 450 \\
uniform re-timing $2\times$ & 96\% @ 27 & 60\%$_{54\text{--}65}$ @ 83 & 64\%$_{59\text{--}68}$ @ 110 & 0\% @ 304 \\
\midrule
within-limit re-timing, $f_{\max}=2$ & 98\% @ 33 [1.40] & 96\% @ 92 [1.25] & 76\%$_{74\text{--}78}$ @ 115 [1.49] & 24\% @ 339 [1.51] \\
within-limit re-timing, $f_{\max}=4$ & 99\% @ 30 [1.57] & 96\%$_{94\text{--}97}$ @ 85 [1.37] & 61\%$_{58\text{--}65}$ @ 101 [1.74] & 23\%$_{18\text{--}30}$ @ 251 [1.89] \\
\bottomrule
\end{tabular}

%% file: tables/controls.tex
\begin{tabular}{lcccc}
\toprule
route & \textsc{Lift} & \textsc{Can} & \textsc{Square} & \textsc{ToolHang} \\
\midrule
within-limit, $f_{\max}=4$ & 99\% @ 30 [1.57] & 96\%$_{94\text{--}97}$ @ 85 [1.37] & 61\%$_{58\text{--}65}$ @ 101 [1.74] & 23\%$_{18\text{--}30}$ @ 251 [1.89] \\
\quad shuffled placement, matched tempo & 99\% @ 34 [1.57] & 84\% @ 96 [1.37] & 59\% @ 119 [1.74] & 10\% @ 428 [1.89] \\
\quad entropy-guided merging (low half) & 94\% @ 34 [1.36] & 87\% @ 93 [1.35] & 76\% @ 136 [1.35] & 12\% @ 380 [1.35] \\
\quad speed-conditioned, best cell & 90\% @ 28 [2.00] & 67\% @ 118 [1.00] & 74\% @ 128 [1.50] & 14\% @ 610 [1.00] \\
\bottomrule
\end{tabular}

%% file: tables/phase.tex
\begin{tabular}{lccccc}
\toprule
surviving exceedances & tempo & $\rho$ & phase split (tr/ap/co, \%) & $\hat\eta$ mean / p90 (mm) & success @ time \\
\midrule
spread over all phases & 1.33 & 0.20 & tr 28, ap 8, co 65 & 11.3 / 27.5 & 90\%$_{89\text{--}91}$ @ 95 \\
none left in transport & 1.33 & 0.20 & tr 0, ap 9, co 91 & 13.1 / 39.0 & 84\%$_{78\text{--}86}$ @ 101 \\
none left in approach & 1.33 & 0.20 & tr 31, ap 0, co 69 & 11.7 / 29.8 & 91\%$_{86\text{--}93}$ @ 95 \\
none left in contact & 1.33 & 0.20 & tr 58, ap 16, co 26 & 9.4 / 22.2 & 92\% @ 92 \\
\bottomrule
\end{tabular}

%% file: tables/general.tex
\begin{tabular}{lccc}
\toprule
setting & nom. & uni. $2\times$ & within-limit $f_{\max}{=}4$ \\
\midrule
\textsc{Can}, flow-matching policy & 96\% & 33\% & 90\%$_{88\text{--}91}$ @ 88 [1.37] \\
\textsc{Square}, flow-matching policy & 85\% & 59\% & 70\%$_{68\text{--}74}$ @ 124 [1.38] \\
\textsc{Can}, mixed-skill demos & 55\% & 66\% & 80\% @ 112 [1.72] \\
\textsc{Lift}, mixed-skill demos & 77\% & 90\% & 96\% @ 37 [2.49] \\
\textsc{Lift}, machine demos (no slack) & 66\% & 30\% & 65\% @ 60 [1.10] \\
\textsc{Can}, 50 demonstrations & 73\% & 21\% & 84\% @ 87 [1.35] \\
\textsc{Can}, chunk 16, execute 8 & 94\% & 6\% & 92\% @ 84 [1.37] \\
\bottomrule
\end{tabular}

%% file: tables/g1.tex
\begin{tabular}{lccc}
\toprule
arm & tempo & success & no fall \\
\midrule
\multicolumn{4}{l}{\emph{uniform time scaling}} \\
\quad $1.25\times$ & 1.25$\times$ & 95\% & 96\% \\
\quad $1.5\times$ & 1.50$\times$ & 82\% & 93\% \\
\quad $2\times$ & 2.00$\times$ & 51\% & 90\% \\
\quad $3\times$ & 3.00$\times$ & 27\% & 87\% \\
\midrule
\multicolumn{4}{l}{\emph{within-envelope re-timing}} \\
\quad $k_{\max}=1.5$ & 1.37$\times$ & 96\% & 98\% \\
\quad $k_{\max}=2$ & 1.65$\times$ & 87\% & 97\% \\
\quad $k_{\max}=4$ & 2.32$\times$ & 71\% & 96\% \\
\midrule
shuffled placement, $k_{\max}=4$ & 2.32$\times$ & 52\% & 92\% \\
\bottomrule
\end{tabular}

%% file: tables/vla.tex
\begin{tabular}{lcccc}
\toprule
route & tempo & $\rho$ & per seed & success @ time \\
\midrule
frozen, nominal speed & 1.00 & 0.00 & 88/88/85 & 87\% @ 114 \\
uniform $2\times$ & 1.99 & 0.59 & 48/49/51 & 49\% @ 86 \\
within-limit, $f_{\max}{=}4$ & 1.39 & 0.00 & 75/72/66 & 71\% @ 87 \\
\bottomrule
\end{tabular}